\pdfoutput=1

\documentclass[11pt]{article}

\usepackage{fancyhdr}
\usepackage[final]{acl}

\usepackage{times}
\usepackage{latexsym}
\usepackage{graphicx}
\usepackage[most]{tcolorbox}
\tcbuselibrary{listings}
\usepackage{booktabs}
\usepackage{colortbl} 
\usepackage{multicol} 

\usepackage{xcolor}

\lstdefinestyle{custombox}{
    frame=single,
    framesep=1pt,
    rulecolor=\color{black},
    numbers=left,
    numberstyle=\small\color{black},
    numbersep=5pt,  
    basicstyle=\fontsize{8.5}{10.5}\selectfont\ttfamily,
    backgroundcolor=\color{gray!10},
    breaklines=true,
    breakindent=0pt,
    xleftmargin=15pt, 
    xrightmargin=10pt,
    framexleftmargin=1pt  
}

\usepackage[T1]{fontenc}

\usepackage[utf8]{inputenc}

\usepackage{microtype}

\usepackage{inconsolata}

\title{QuaLLM: An LLM-based Framework to Extract Quantitative Insights from Online Forums}

\author{Varun Nagaraj Rao \\
  Princeton University / CITP \\
  \texttt{varunrao@princeton.edu} \\\And
  Eesha Agarwal \\
  Princeton University \\
  \texttt{eagarwal@alumni.princeton.edu} \\\AND
  Samantha Dalal \\
  University of Colorado, Boulder \\
  \texttt{samantha.dalal@colorado.edu} \\\And
  Dana Calacci \\
  Penn State University\\
  \texttt{dcalacci@psu.edu} \\\And
  Andrés Monroy-Hernández \\
  Princeton University / CITP\\
  \texttt{andresmh@princeton.edu} \\  
  }

\begin{document}
\maketitle

\begin{abstract}
Online discussion forums provide crucial data to understand the concerns of a wide range of real-world communities. However, the typical qualitative and quantitative methodologies used to analyze those data, such as thematic analysis and topic modeling, are infeasible to scale or require significant human effort to translate outputs to human readable forms. This study introduces QuaLLM, a novel LLM-based framework to analyze and extract quantitative insights from text data on online forums. The framework consists of a novel prompting and human evaluation methodology. We applied this framework to analyze over one million comments from two of Reddit's rideshare worker communities, marking the largest study of its type. We uncover significant worker concerns regarding AI and algorithmic platform decisions, responding to regulatory calls about worker insights. In short, our work sets a new precedent for AI-assisted quantitative data analysis to surface concerns from online forums. 
\end{abstract}

\section{Introduction}

Text data from online forums, such as Reddit, has been used to study issues affecting diverse communities~\cite{proferes2021studying}. Researchers have leveraged this data across various domains, including healthcare~\cite{zamanifard2023social,xu2023technology, ho2017stigma, paxman2021everyone}, political discourse~\cite{papakyriakopoulos2023upvotes, de2021no, marchal2020polarizing}, education~\cite{madsen2022communication, park2018harnessing}, and gig work\footnote{A market characterized by short-term and freelance work like food delivery and ridesharing}~\cite{rosenblat2016algorithmic, sannon2022privacy, yao2021together, watkins2022have, ma2018using}, to highlight critical concerns and trends. 

\begin{figure}[htb!]
    \centering
    \includegraphics[width=\linewidth]{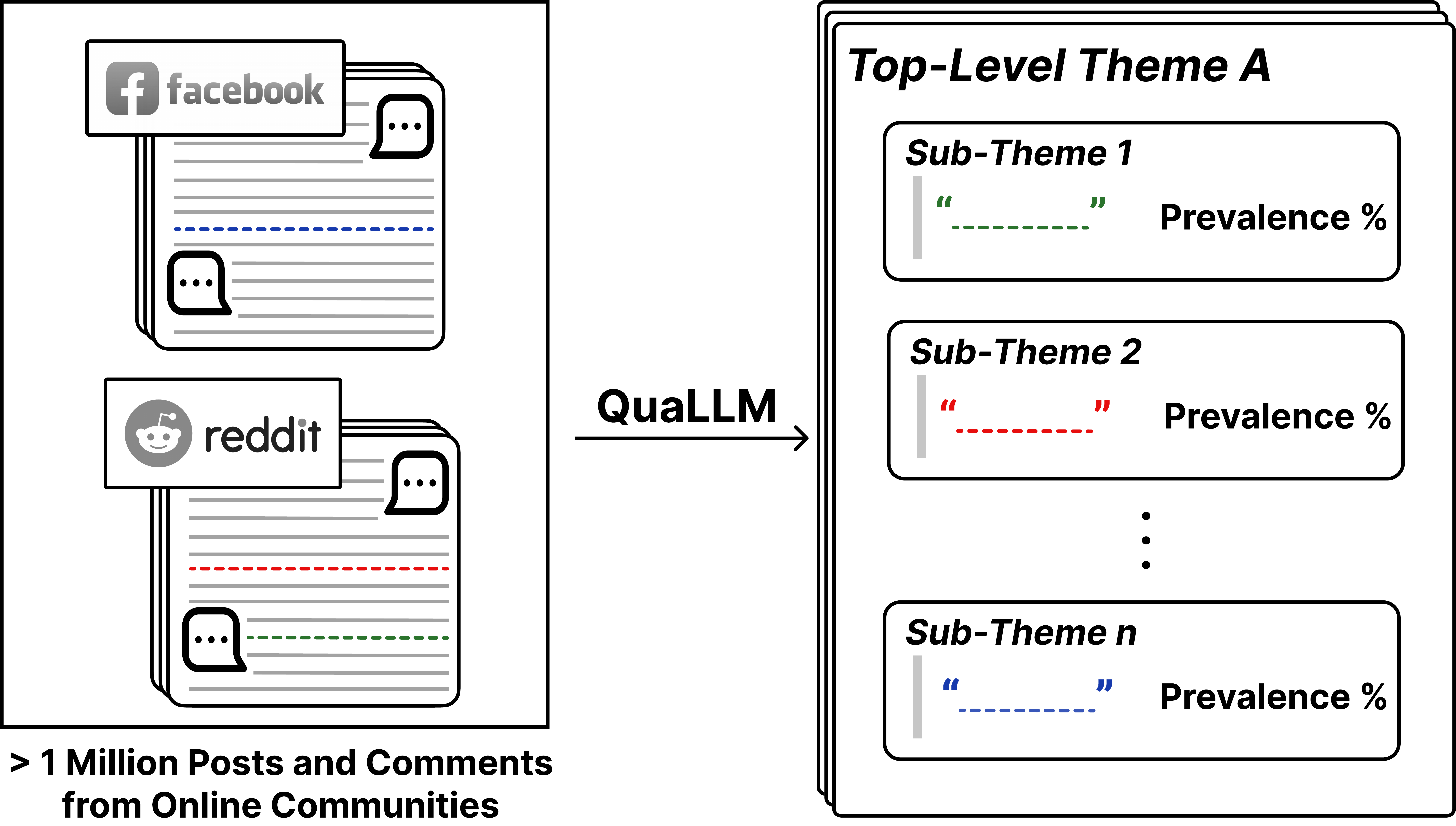}
    \caption{QuaLLM transforms large-scale unstructured text-based online forum discussions on platforms like Reddit and Facebook into a structured survey-style format, identifying top-level themes associated with prevalence ranked sub-themes (by frequency of occurrence) and representative quotes.}
    \label{fig:teaser_figure}
\end{figure}

\begin{figure*}[htb]
    \centering
    \includegraphics[width=\textwidth]{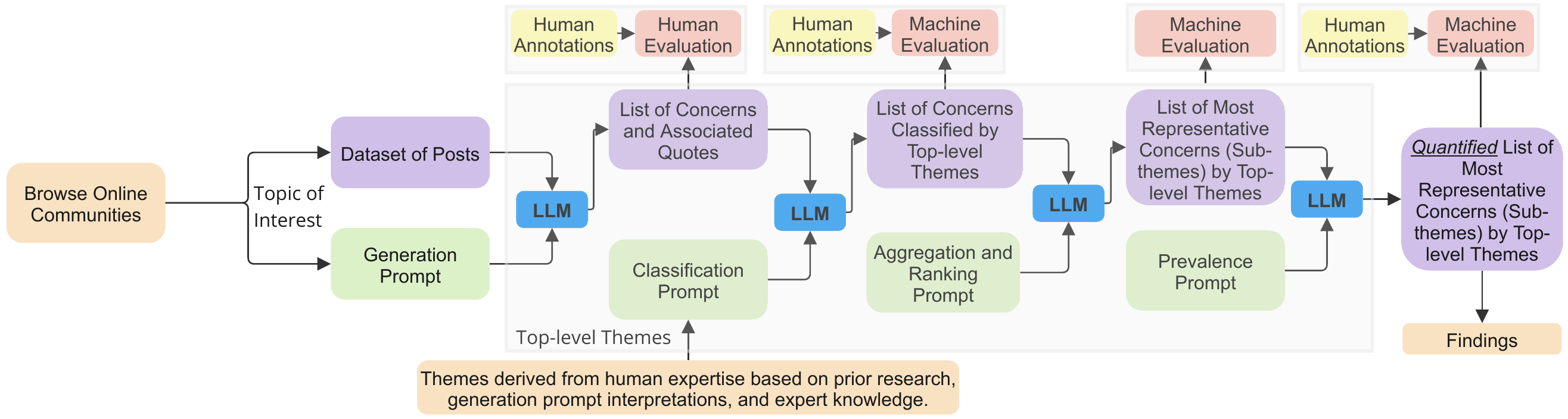}
    \caption{QuaLLM's multiphase prompting attempts to evoke the steps that human analysts might perform.}
    \label{fig:methods-overview}
\end{figure*}
Studies analyzing this text data for social phenomena often use interpretive qualitative methodologies~\cite{gichuru2017interpretive} like coding~\cite{saldana2021coding} and thematic analysis~\cite{braun2012thematic}, to extract themes. Although detailed, these methodologies demand considerable human effort and interpretation. Such detailed interpretations may not always be necessary for identifying broad trends across large online communities. Alternatively, positivist quantitative~\cite{park2020positivism} techniques like LDA~\cite{abdul2018trends} and BERT derivatives for topic modeling distill topics, which humans can interpret and convert into themes~\footnote{We distinguish between \textit{topics/concepts} and \textit{themes}; a \textit{topic/concept} is a concrete and specific subject matter of a text often expressed as a single word or a short phrase, while a \textit{theme} is an interpretive and abstract insight conveyed by the text, typically articulated as a complete sentence.}, offering greater scalability. However, they require careful hyperparameter selection, operate on limited context length, and require significant human effort to translate outputs to human readable forms.  

Recent advancements in LLMs and prompt engineering~\cite{shah2023using,nori2023can, openai2023prompt, deng2023llms} offer an alternative for analyzing extensive text data from online forums, addressing limitations of scale and context. Specifically, LLMs have been shown to perform well on generation, classification, and ranking tasks~\cite{zhao2023survey}, and thus have the potential to assist researchers in text data analysis. However, the lack of an effective methodology for extracting themes from online communities limits their use. We address this with the following contributions: 
\begin{enumerate}
    \itemsep-0.5em
    \item[(i)] We introduce QuaLLM, an LLM-based framework consisting of a novel prompting and human evaluation methodology for the thematic analysis and extraction of quantitative insights from online forums' text data (See Figure \ref{fig:teaser_figure}).
    \item[(ii)] We apply our framework to a case study on Reddit's rideshare communities, analyzing over one million comments—the largest study of its kind—to identify worker concerns regarding AI and algorithmic platform decisions, responding to regulatory calls~\cite{fed2023worker}.
    \item[(iii)] We discuss the broader implications of using LLMs for quantitative text data analysis.
\end{enumerate}
Taken together, our work establishes a new precedent for AI-assisted quantitative data analysis to surface concerns from online forums.

\section{Related Work}

\noindent \textbf{LLM Prompting and Evaluation.} \qquad 
LLMs have been used for generation of topics and concepts~\cite{shah2023using,zhao2023survey}\footnote{The novelty of our work lies in surfacing themes (e.g., ``facing fear leads to growth'') rather than topics or concepts (e.g., ``fear,'' ``love,'' ``bias of LLMs'')}, classification~\cite{deng2023llms}, and ranking~\cite{qin2023large} tasks via advances in prompt engineering~\cite{nori2023can, openai2023prompt}. Popular prompting methods include: \textit{In-context learning}~\cite{brown2020language}, which teaches LLMs new tasks with few labeled examples, \textit{Chain of Thought}~\cite{wei2022chain}, which breaks down complex problems through stepwise reasoning, and \textit{Ensembling},~\cite{wang2022self} which aggregates outputs of multiple model runs to obtain a consensus output. Despite these advances, accurately evaluating LLM outputs is challenging, with human evaluation generally preferred~\cite{cohere2023eval}. Building on prior work, we introduce a framework consisting of a novel prompting and human evaluation methodology for the quantitative analysis of text data from online forums.
\vspace{0.5em}

\noindent \textbf{Gig Work and Reddit Analysis.} \qquad 
Regulatory bodies~\cite{fed2023worker} have sought public insights on the impact of AI and algorithmic management by employers~\cite{nytimes2022worker, bernhardt2021work}, particularly in the gig economy. Gig workers actively discuss their experiences on anonymous forums, e.g., Reddit, which have become a rich research data source. Previous studies on rideshare worker concerns using Reddit, highlighting issues of privacy, scams, and support systems~\cite{rosenblat2016algorithmic, sannon2022privacy, yao2021together, watkins2022have, ma2018using}, have been small-scale, limited by manual coding\footnote{the largest dataset analyzed about 2.6K posts~\cite{watkins2022have} using qualitative data analysis methodologies}. We use QuaLLM to examine over 1 million comments from rideshare workers, significantly broadening the scope to amplify all voices and also to translate this academic research into policy-relevant insights~\cite{nagarajrao2024memo, nagarajrao2024gpai, nagarajrao2024rideshare}. 
\section{QuaLLM Framework}
We introduce QuaLLM (Figure \ref{fig:methods-overview}), a framework consisting of a prompting method and evaluation strategy for the thematic analysis of text data from online forums. 

\subsection{LLM Prompting Method for Analysis of Online Forum Data}
Our approach begins with collecting datasets from specified periods across any public threaded discussion forum, such as Reddit and Facebook. We aim to extract and summarize the predominant concerns of workers within these communities through a four-step prompting process: generation, classification, aggregation, and prevalence.

\noindent \textbf{Generation Prompt.} \qquad
The generation step involves the generation of \textit{concern summaries} from online discussions \textit{relevant to a specific topic} and linking each concern with a \textit{representative quote}. The prompt is broken down into seven steps, following best prompt engineering practices from prior research and industry~\cite{openai2023prompt, adams2023sparse, nori2023can, wang2023grammar}. The steps include: 
\begin{enumerate}
\itemsep-0.5em 
    \item identifying concerns
    \item aggregating to avoid redundancy
    \item selecting representative quotes\footnote{we argue that forcing the model to choose and output quotes is a form of Chain-of-Thought prompting~\cite{wei2022chain} strategy and find that it helps prevent hallucination}
    \item assessing concern frequency and impact
    \item formatting outputs in JSON with concern titles, descriptions, and quotes
    \item ensuring concerns are topic-specific and derived from input data without external knowledge
    \item reiterating to prevent redundancy and indicating when no concerns are present
\end{enumerate}
To minimize API calls, we aggregate posts and comments into groups. For each group, we programmatically enhance the JSON output with the timestamp of the earliest post and assign a unique identifier.
All these steps are necessary since direct prompting may cause hallucinations, redundancy, and inconsistent output.
This prompting strategy is similar to ``open coding,'' where one derives new theories and concepts informed from the underlying data~\cite{saldana2021coding}.

\noindent \textbf{Classification Prompt.} \qquad
We define a set of top-level themes, typically around 4-5, based on prior research, parallel participant interviews, or other expert input, and the outputs of the generation prompt. We then prompt the LLM to classify each identified concern accordingly, with an additional ``other'' category for irrelevant concerns. This step mirrors ``thematic analysis''~\cite{braun2012thematic}, which groups together a number of related codes, serving as a filtration mechanism.

\noindent \textbf{Aggregation Prompt.} \qquad
Aggregates concerns into sub-themes based on classifications and identifies representative concerns (e.g., top 5) within each top-level theme. We manually select representative quotes for each sub-theme.

\noindent \textbf{Prevalence Prompt.} \qquad
Classifies concerns within top-level themes as belonging to one of the sub-themes or an ``other'' category, facilitating quantitative analysis of concern prevalence by theme percentage. 

\subsection{LLM Outputs Evaluation Strategy}
To validate the LLM outputs, we propose a mix of human and computational evaluation methods based on the following metrics: \textit{factuality} and \textit{completeness} for generation, \textit{accuracy} for classification and prevalence, and \textit{distinctness} and \textit{coverage} for aggregation prompts, respectively. 

\subsubsection{Generation Prompt:} This prompt generates new text given unstructured source text as input.

\noindent \textbf{Factuality:}  Determines if the LLM's output concerns reflect the source data. Evaluators respond to: \textit{``Is this candidate concern (and quote) output by the LLM present in the reference human generated list of transparency concerns? Answer Yes or No''}. The factuality score is the proportion of accurate LLM-generated concerns akin to precision.

\noindent \textbf{Completeness:}  Evaluates the LLM's ability to capture all relevant concerns, asking: \textit{``Is this reference human concern present in the candidate concern (and associated quote) output by the LLM? Answer Yes or No.''} The completeness score denotes the proportion of human-identified concerns recognized by the LLM, akin to recall.

\subsubsection{Classification and Prevalence Prompts:} Both these prompts classify source text into predefined categories.

\noindent \textbf{Accuracy:} Assesses the correctness of LLM classifications, measuring the proportion of accurately classified samples.

\subsubsection{Aggregation Prompt:} This prompt generates aggregated text given source text as input. We leverage a topic model to generate topics for the entire text and find those topics most similar to the subthemes. Human evaluation on a subset does not accurately capture the task, and instead, we pivoted to a computational evaluation of the entire text using well-established topic modeling techniques.

\noindent \textbf{Distinctness:}  We measure the \textit{proportion of unique most-similar topics associated with each sub-theme across all the sub-themes}. A higher value indicates a greater diversity in the subthemes. 

\noindent \textbf{Coverage (k):} We measure the proportion of unique most-similar topics associated with each of the `n' sub-themes that are also found among the most frequent `nk' topics of the overall text. For example, if n=5 and k=2, we would look at the 5 sub-themes amongst the top 10 most frequent overall topics.
A higher CR-k implies the sub-themes represent prevalent or significant themes in the text.
\section{Case Study of Worker Concerns from Rideshare Subreddits}

We apply our framework to rideshare subreddit data (\texttt{r/uberdrivers, r/lyftdrivers}) using the GPT4-Turbo model on Microsoft Azure to identify worker concerns regarding AI and platform algorithms. See Appendices \ref{asec:prompts}, \ref{asec:findings}, and \ref{asec:costs} for our prompts, findings, and costs incurred, respectively. 

We chose rideshare subreddits for their active 400K+ member community, presence of active discussions of worker concerns, which is of interest to regulators, accessible and comprehensive data, and relevance to Human-Computer Interaction, Economics, and Communications researchers studying the harms caused to workers as a result of rideshare platforms' algorithmic management~\cite{nagarajrao2024rideshare,zhang2022algorithmic, dubal2023algorithmic, rosenblat2016algorithmic}. 
We now describe our dataset and summarize our findings.

\subsection{Dataset and Experiments}
We obtained comprehensive 2019-2022 data from \texttt{r/uberdrivers} and \texttt{r/lyftdrivers} via \url{https://the-eye.eu/redarcs/}, focusing on this period to capture recent concerns, particularly relevant due to app changes and the unavailability of post-2022 data due to Reddit API changes\footnote{Reddit API restrictions enforced in Feb'23; see \url{https://www.redditinc.com/policies/data-api-terms}}. After processing to remove short samples, our dataset\footnote{On Reddit, a submission is a post, and a comment is a response to a post or another comment. We concatenate and use the text from posts and all associated comments} comprised 65,377 submissions  ($47,106$ from Uber, $18,271$ from Lyft) and $1,392,776$ associated comments ($1,054,030$ from Uber, $338,746$ from Lyft). We then grouped submissions and comments into sets of five for each LLM API call (See Appendix \ref{asec:prompts} for the prompts). We encountered an 11\% error rate due to throttling and content issues. Finally, we obtained 58,728 concerns which we categorized into four themes: Enhancing Transparency and Explainability (24,721 concerns, 42\%), Predictability and Worker Agency (12,728 concerns, 22\%), Better Safety and More Time (6,144 concerns, 10.5\%), and Ensuring Fairness and Non-Discrimination (4,280 concerns, 7\%); the remaining 18.5\% were categorized as ``Other''. We then aggregated these findings to highlight the top five concerns for each theme and calculated their prevalence across the dataset.

\subsection{Findings}
We briefly summarize our findings below. We also provide tables of the aggregated, ranked, and prevalence-quantified themes in Appendix \ref{asec:findings}. 

\noindent \textit{(i) Enhancing Transparency and Explainability:} Drivers are concerned about opaque fare calculations, unclear incentives, and uncertain criteria for earnings, surge pricing, and cancellations (Table \ref{tab:reddit-transparency})

\noindent \textit{(ii) Predictability and Worker Agency:} Drivers face unpredictable earnings from fluctuating surge pricing, algorithm changes, increased competition, complex incentive qualifications, and low compensation for long pickups and waits. (Table \ref{tab:reddit-predictability})

\noindent \textit{(iii) Better Safety and More Time:} Drivers face navigation issues, support access difficulties,  challenges with false complaints, and low compensation for additional tasks and wait times. (Table \ref{tab:reddit-safety})

\noindent \textit{(iv) Ensuring Fairness and Non-Discrimination:} Drivers dispute their fare share and platform take rates, fairness in ride assignments, the influence of demographics on ratings, and the criteria for deactivation. (Table \ref{tab:reddit-fairness})

\subsection{Evaluation}
\begin{table*}[htb]
\resizebox{\textwidth}{!}{%
\begin{tabular}{@{}l|lllll@{}}
\toprule
\rowcolor{gray}
\color{white}\textbf{Evaluation Metrics} &
  \multicolumn{1}{c}{\cellcolor{gray}\color{white}\textbf{Factuality}} &
  \multicolumn{1}{c}{\cellcolor{gray}\color{white}\textbf{Completeness}} &
  \multicolumn{1}{c}{\cellcolor{gray}\color{white}\textbf{Distinctness}} &
  \multicolumn{1}{c}{\cellcolor{gray}\color{white}\textbf{Coverage (k)}} &
  \multicolumn{1}{c}{\cellcolor{gray}\color{white}\textbf{Accuracy}} \\ \midrule
\textbf{\begin{tabular}[c]{@{}l@{}}Analysis phase \\ evaluated with \\metric\end{tabular}} &
  Generation &
  Generation &
  Aggregation &
  Aggregation &
  \begin{tabular}[c]{@{}l@{}}Classification, \\ Prevalence\end{tabular} \\ \midrule
\textbf{\begin{tabular}[c]{@{}l@{}}Reference data\\ generation \\ method\end{tabular}} &
  \begin{tabular}[c]{@{}l@{}}Theme extraction \\ by human \\ (2 researchers)\end{tabular} &
  \begin{tabular}[c]{@{}l@{}}Theme extraction \\ by human \\ (2 researchers)\end{tabular} &
  \begin{tabular}[c]{@{}l@{}}Topic extraction using \\ topic models \\ (BERTopic)\end{tabular} &
  \begin{tabular}[c]{@{}l@{}}Topic extraction using \\ topic models \\ (BERTopic)\end{tabular} &
  \begin{tabular}[c]{@{}l@{}}Labeling \\ by humans \\ (2 researchers)\end{tabular} \\ \midrule
\textbf{\begin{tabular}[c]{@{}l@{}}Comparison b/w \\ reference data and \\ LLM output\end{tabular}} &
  \begin{tabular}[c]{@{}l@{}}Manually \\ (2 researchers)\end{tabular} &
  \begin{tabular}[c]{@{}l@{}}Manually \\ (2 researchers)\end{tabular} &
  \begin{tabular}[c]{@{}l@{}}Automatically \\ (Python Program)\end{tabular} &
  \begin{tabular}[c]{@{}l@{}}Automatically \\ (Python Program)\end{tabular} &
  \begin{tabular}[c]{@{}l@{}}Automatically \\ (Python Program)\end{tabular} \\ \midrule
\textbf{\begin{tabular}[c]{@{}l@{}}Limitation of \\ reference data \\ generation method\end{tabular}} &
  Costly &
  Costly &
  \begin{tabular}[c]{@{}l@{}}Need for interpretation \\ of topic model outputs\end{tabular} &
  \begin{tabular}[c]{@{}l@{}}Need for interpretation \\ of topic model outputs\end{tabular} &
  Costly \\ \midrule
\textbf{\begin{tabular}[c]{@{}l@{}}Evaluation \\ sample size\end{tabular}} &
  \begin{tabular}[c]{@{}l@{}}125 posts\\ 2,511 comments\end{tabular} &
  \begin{tabular}[c]{@{}l@{}}125 posts\\ 2,511 comments\end{tabular} &
  47,873 concerns &
  47,873 concerns &
  100 concerns \\ \midrule
\textbf{Actual data size} &
  \begin{tabular}[c]{@{}l@{}}65,377 posts\\ 1,392,776 comments\end{tabular} &
  \begin{tabular}[c]{@{}l@{}}65,377 posts\\ 1,392,776 comments\end{tabular} &
  47,873 concerns &
  47,873 concerns &
  \begin{tabular}[c]{@{}l@{}}58,728 concerns \\ (classification)\\ 47,873 concerns \\ (prevalence)\end{tabular} \\ \midrule
\textbf{Results} &
  0.55* &
  0.78* &
  0.80* &
  \begin{tabular}[c]{@{}l@{}}0.95* (k=1)\\ 1.00* (k=2)\end{tabular} &
  \begin{tabular}[c]{@{}l@{}}0.74* (classification)\\ 0.82* (prevalence)\end{tabular} \\ \bottomrule
\end{tabular}%
}
\caption{Metrics Chosen to Evaluate LLM-Powered Analysis Phases and Results. Metric values belong to the closed interval {[}0,1{]}, with higher values indicative of better performance.  * values are statistically significantly different from chance measured via a binomial test (p-value \textless 0.05). 					}
\label{tab:llm-evaluation}
\end{table*}

\subsubsection{Generation Prompt.}  We randomly sampled 125 submissions and all the 2,511 associated comments, jointly annotated by two researchers, to create a reference list of concerns. The two researchers then jointly evaluated the LLM outputs against the reference list of concerns and resolved any differences through discussions.
We obtained a factuality score of 0.55 and a completeness score of 0.78. Lower factuality was primarily due to the LLM identifying concerns that are part of the input context but not directly related to AI and algorithmic decisions. Moreover, the lower factuality score was less concerning as outputs that are not factual (i.e., identified by the LLM but not by humans) were mostly filtered out into the ``Other'' category during the classification step and excluded from any further analysis. 

\subsubsection{Classification and Prevalence Prompts.}  Two researchers jointly annotated 100 LLM-classified concerns to establish ground truth labels, assigning one label per concern. They compared LLM outputs with these labels, achieving a consensus accuracy of 0.74. The achieved accuracy is significant because misclassified samples, excluding those labeled as ``Other,'' can still contribute to the analysis. This is because concerns might fit into multiple categories depending on the analytical perspective. These concerns remain valuable for later aggregation and prevalence prompts, even if placed in a different category.
Using a similar method, researchers obtained an accuracy of 0.82 for the prevalence prompts' output.

\subsubsection{Aggregation Prompt.} We employed BERTopic for topic modeling (see Appendix \ref{asec:bertopic-hyperparams} for hyperparameter details). Given the large volume of concerns (4K-25K) per top-level theme, human evaluation on a subset would not accurately capture the task's nuances. Therefore, we adopted this well-established computational method for evaluation. We fit a topic model on the entire text (title+description) associated with each top-level theme to obtain the top 5 and 10 topics. Subsequently, we identified the most similar topic for each of the 5 sub-themes. The results demonstrated a distinctness of 0.80, coverage(1) of 0.95, and coverage(2) of 1.00, indicating that the LLMs' outputs were distinct and well-represented among the most frequent topics.  

\section{Conclusion}

Our study introduces a new LLM-based framework with a prompting method and evaluation strategy to analyze online forum data, demonstrated through the largest analysis of over 1 million rideshare subreddit comments to date. The methodology's recursive nature demonstrates its versatility, e.g., the Aggregation and Prevalence prompts can function as Generation and Classification at the sub-theme level, and initial filtering can reframe as classifying concerns' relevance to the overall theme. This approach highlights LLMs' capability to efficiently process large unstructured datasets, establishing a precedent for future AI-assisted positivist quantitative research. But does an increase in data equate to enhanced understanding? Echoing \citet{agnew2024illusion}'s sentiments, we believe further research is essential.
\section*{Ethical and Broader Impacts Statement}
\label{sec:ethics}
\subsection*{Scaling online forum analysis with LLMs}
QuaLLM presents a novel framework that leverages the power of LLMs to analyze vast amounts of unstructured text data efficiently and translate it to a form that resembles community survey responses. Beyond analyzing online discussions in rideshare communities on Reddit, this methodology has broad applicability across numerous domains, across any platform, where large amounts of unstructured text are prevalent and traditional positivist quantitative methodologies~\cite{park2020positivism} such as topic models through LDA, may fall short. 
\begin{description}
    \itemsep-0.5em
    \item [Healthcare Forums:] Identify discussions of medications from patient forums comparing anecdotal experiences and widespread concerns.
    \item [Customer Feedback:] Filter through customer service chats and emails to identify recurring product or service issues.
    \item [Policy and Regulation:] Review public responses to policy proposals, identifying key points of support and contention.
    \item [Educational Feedback:] Review online forums and course feedback to extract themes related to assessment and course content.
    \item [Financial Market Sentiments:] Analyze investor forums and financial news comments to gauge market sentiment towards certain stocks or economic policies.
\end{description}

\subsection*{QuaLLM's evaluation reveals strengths and necessitates further refinement}
Due to the unique nature of our pipeline approach, which combines several tasks into a cohesive workflow, there are no established baselines or benchmarks available for direct end-to-end comparison and evaluation. Topic models might offer a comparison for the generation prompt outputs, yet they require significant human effort to translate outputs to human-readable forms, unlike our methodology, where the LLM autonomously generates topics, making baselining challenging.

This absence of precedent opens questions regarding effective baselining and benchmarking for performance evaluation. 
As a result, human evaluation serves as the gold standard for evaluation.

\subsection*{LLMs can be used to assist researchers and complement existing methods}

Some researchers highlight limitations in LLMs' understanding of complex human emotions and experiences, pointing to a lack of context-sensitivity~\cite{bender2021dangers, alkaissi2023artificial, rudolph2023chatgpt}. Others~\cite{bano2023exploring} argue LLMs cannot match the innate human ability to perform specific tasks in inexplicable ways, described in conceptual frameworks for knowledge creation~\cite{li2003nonaka, collins2005tacit}. This gap raises doubts about LLMs' capability for interpretation crucial to fields such as anthropology~\cite{malinowski1929practical} and sociology~\cite{weber1949objectivity}, challenging their application within the qualitative interpretivist research paradigm ~\cite{gichuru2017interpretive, glaser1968discovery, denzin2001interpretive}. 

On the other hand, advocates for LLM-enhanced research suggest LLMs align with human outputs and may replace human judgement~\cite{dwivedi2023so, bano2023exploring, chew2023llm, xiao2023supporting, dai2023llm, tai2023examination, dunivin2024scalable}. Further, they find integrating AI with human expertise can improve the scope and scale of analyses, facilitating efficient processing of vast, unstructured datasets. 

Our research stands in the middle and underscores the potential of LLMs to help researchers engage with data in a way that can supply intrinsic meaning at scale while stressing the need to recognize their limitations and the continued importance of human oversight and interpretation. 

Broadly, QuaLLM can also be used in mixed methods study designs to complement data from interviews and focus groups with adequate safeguards and evaluations.

\section*{Limitations}
\begin{description}
    \itemsep-0.5em
    \item[Interpretivist vs. Positivist trade-offs:] While LLMs efficiently aggregate broad trends, reflecting a positivist quantitative approach that assumes text has intrinsic meaning, individual experiences may be neglected. This contrasts with the interpretivist qualitative paradigm, which values subjective meaning derived from the interaction between the researcher and the data.
    \item [Small scale human evaluation:] We acknowledge the limited generalizability of our human evaluation due to the relatively small sample size 
    \item [Hyperparameter-sensitive topic modeling:] BERTopic performance is very sensitive to the choice of hyperparameters, and this can result in significant variation in results based on the chosen hyperparameter values.
    \item [Reproducibility concerns:] The stochasticity of LLM outputs raises concerns about reproducibility. However, this variability mirrors the inherent subjectivity of human perspectives, which can differ across time and contexts.
    \item [Bias in LLMs:] LLM outputs are influenced by biases introduced during training, alignment processes, and prompt design, which may affect what concerns are highlighted, favor structured language over informal text, and introduce potential socio-economic or stylistic biases; additionally, API errors (e.g., due to profanity) may create selection bias in the analyzed data.
    \item [Misinformation in Online Communities:] Misinformation in online forums can distort analysis, and the spread of false information may reflect user biases or socio-economic factors. However, the noise may smoothen out when aggregated over millions of posts, reducing its overall impact.
\end{description} 

\section*{Acknowledgements}
We acknowledge the Azure Cloud Computing Grant from CSML at Princeton, which enabled access to Microsoft Azure compute resources. We thank Sayash Kapoor for insightful discussions on the conceptual and normative implications of LLMs and prompting strategies. We are grateful to Vinayshekhar Bannihatti Kumar for his valuable input on defining evaluation metrics. We thank Sunnie Kim for her contributions on connections to Explainable AI. We appreciate Elizabeth Anne Watkins for providing valuable critiques from an interpretivist perspective, which helped reframe our work. We thank Yuhan Liu, Beza Desta, and Kiyosu Maeda from the Princeton HCI group for their annotations. We are grateful to Hope Schroeder for her feedback on evaluations and broad discussions on applications. We thank Prof. Arvind Narayanan and Mihir Kshirsagar from Princeton CITP for their insights on the generalization of this work and its policy implications. We thank Prof. Brian Keegan for pointing us to the Reddit data source. Finally, we thank the anonymous CSCW and FAccT reviewers for their valuable feedback, which helped frame the larger paper of which this is a part.

\bibliography{references}
\appendix
\onecolumn
\section{Prompts}
\label{asec:prompts}
\begin{lstlisting}[style=custombox, caption={GPT-4 Generation Prompt}]
Analyze a set of JSON objects, each representing a submission from the r/UberDrivers and r/LyftDrivers subreddits. For each JSON object, you will find the following information:

Submission Title: The title of the Reddit post.
Submission Body: The main content or message of the post.
Timestamp: The date and time when the post was submitted.
Group Key: A unique identifier that is common across all 5 submissions in the dataset.
Comments: A list of all comments made on the submission.

Generate a list of the most frequently occurring and impactful concerns due to AI and algorithmic platform features discussed by drivers within the input context.

Step 1: Identify mentions in submission bodies, titles, and/or comments about concerns that pertain to a lack of knowledge or available information regarding the platforms' algorithms for drivers, including but not limited to fares, routes, incentive programs, driver preferences, etc.
 
Step 2: Group similar concerns across comments and submissions to ensure a mutually exclusive list of concerns and avoid redundancy. For example, multiple mentions of fare calculation issues should be grouped under a single concern.

Step 3: From the grouped concerns, select the most representative quote for each concern. Ensure the quote clearly illustrates the specific concern due to AI and algorithmic platform features.

Step 4: Assess which concerns are mentioned most frequently and have the most significant impact on drivers.

Step 5: Create a list of these concerns in a JSON format. Each entry should include (with these specific field names):
"title": The title of the concern
"description": A brief description (10-20 words)
"quote": The selected representative quote.

Step 6: Ensure the final list is concise, precise, and specifically addresses drivers' concerns due to AI and algorithmic platform features present regarding the platform's algorithms and policies. Include only those concerns found in the input context without generalizing based on prior or outside knowledge. 

Step 7: Group any similar concerns to avoid redundancy. If there are no concerns, output "No concerns". Do not generate any other text.
\end{lstlisting}

\clearpage
\begin{lstlisting}[style=custombox, caption={GPT-4 Classification Prompt}]
Task: Analyze a list of rideshare drivers' transparency concerns. Each concern should be evaluated and categorized into one of the following five categories. For each concern listed, assign exactly one letter that corresponds to its most appropriate classification in the same order as the original list (preserving the serial number of the concern).

Categories:

A. Need for Enhancing Transparency and Explainability: Concerns due to the lack of information and explanations about AI and algorithmic features, required for drivers to do their work. These include opaque trip details, unclear surge boundaries, bonus and quest clarity issues, lack of wage breakdown, platform take rate and clarity on how prices and bonuses are calculated, and the impact of ratings on metrics.

B. Need for Greater Predictability and Worker Agency: Concerns due to significant variation in AI and algorithmic features leading to reduced worker agency and diminished predictability of work conditions. These include unpredictable wages, misleading destination filter, immense variation in surge prices and quest matches across drivers and location, and deceptive trip offers.

C. Need for Better Safety and More Time: Concerns focused on safety risks and time pressures due to the AI and algorithmic features. These include dangerous multitasking, compromised route safety, unattainable quests, and acceptance rate concerns, emphasizing the need for improved safety and time management in app design.

D. Need for Ensuring Fairness and Non-Discrimination: Concerns related to algorithmic wage discrimination. These include unequal pay for similar work, earnings below the prevailing minimum wage, influence of the demographic characteristics of drivers and riders on earnings, all caused by opaque AI and algorithmic features.

E. Other Concerns: Any concerns that do not fit into the above categories.

Output Format: Present the analysis in a dictionary format with the serial number of the concern as the key and the classification as the value, preserving the order of the concerns.

{1: A, 2: B, 3: C}

Note: Ensure that each concern is classified under only one of the aforementioned categories and that there is one classification corresponding to each concern in the input, e.g., for 400 transparency concerns, there should be 400 items in the dictionary. Do not generate any other text. 
\end{lstlisting}

\clearpage
\begin{lstlisting}[style=custombox, caption={GPT-4 Aggregation and Ranking Prompt}]
The data contains a list of concerns of rideshare drivers obtained from discussions on the r/uberdrivers and r/lyftdrivers subreddits. Each concern title and its description is present on consecutive lines.

These concerns are relevant to <insert category description from Classification Prompt>

Identify the 5 most frequently occurring themes of concerns. Ensure each concern is sufficiently different from the others on the list. Don't repeat the same concern in the list. If there are similar concerns, group them and find another one. 

Provide the output in a rank ordered format:
{concern_rank: <a number between 1-5>, concern_title: string, concern_description: string of 10-20 words}

\end{lstlisting}

\begin{lstlisting}[style=custombox, caption={GPT-4 Prevalence Prompt}]
Classify each line which contains the title and description of a concern from rideshare drivers on Reddit into the following 6 categories:

<list the 5 categories A to E from the aggregation prompt output>
F: Other

Output Format: Present the analysis in a dictionary format with the serial number of the concern as the key and the classification as the value, preserving the order of the concerns.

{1: A, 2: B, 3: C}

Note: Ensure that each concern is classified under only one of the aforementioned categories and that there is one classification corresponding to each concern in the input, e.g., for 400 concerns, there should be 400 items in the dictionary. Do not generate any other text.
\end{lstlisting}

\section{Findings}
\label{asec:findings}

\begin{table}[htb]
\small
\resizebox{\textwidth}{!}{%
\begin{tabular}{@{}p{0.37\linewidth} |p{0.58\linewidth} |p{0.05\linewidth} @{}}
\toprule
\multicolumn{1}{c|}{\textbf{Harm}} &
  \multicolumn{1}{c|}{\textbf{Quote}} & \multicolumn{1}{c}{\textbf{\% (Count)}}\\ \midrule
Drivers are unsure about opaque fare calculation methods. &
  \textit{``\$30 to the middle of nowhere and had to eat dead miles to drive back''} & 29 (7,202)\\ \hline
  Issues with understanding and qualifying for incentive programs, bonuses, and quests. &
  \textit{``Are they really expecting a thumbnail of a map with zero details of where I'd need to be to have a ride qualify for the Quests to be useful?''} & 20 (4,880) \\ \hline
Drivers express concerns over clarity in ride assignments and earnings determination by algorithms. &
  \textit{``A driver who keeps a high AR, low cancel rate, gets mostly 5 Stars, and gives out mostly 5 Stars to pax will get better (more lucrative) rides than a driver the company's algorithm doesn't like.''} & 11 (2,765) \\ \hline
Confusion persists over surge pricing determination and boundaries. &
  \textit{``I get a ton of riders telling me they are paying $40-$60 for the ride (surging rate), but when the ride ends, I get around \$6 (non-surge rate pay)''} & 8 (1,858) \\ \hline
Concerns over lack of clarity in cancellation criteria and the associated fare impact. &
  \textit{``Then I ask about the cancellation fee and they said they can’t provide one as no option came up.''} & 5 (1,197) \\ \bottomrule
\end{tabular}%
}
\caption{Most representative concerns on Reddit relevant to the need for enhancing transparency and explainability (Total = 24,721).}
\label{tab:reddit-transparency}
\end{table}
\begin{table}[htb]
\small
\resizebox{\textwidth}{!}{%
\begin{tabular}{@{}p{0.42\linewidth} |p{0.53\linewidth} | p{0.04\linewidth} @{}}
\toprule
\multicolumn{1}{c|}{\textbf{Harm}} &
  \multicolumn{1}{c|}{\textbf{Quote}} & \multicolumn{1}{c}{\textbf{\% (Count)}}\\ \midrule
Drivers face uncertain and fluctuating income due to changes in surge and quest pricing. &
  \textit{``Uber has always done that. When they offer a higher promo, they manipulate and lower surges.''} & 23 (2,953) \\ \midrule
Challenges with understanding and adapting to algorithm changes that affect ride assignments and income. &
  \textit{``I multi-app, but the algorithm has changed. If you log off from the app, to ride for another app, the algorithm puts you on the bottom of the queue after you log on again for the next ride.''} & 19 (2,408)\\ \midrule
Complexity and unpredictability in qualifying for incentive programs like quests and bonuses. &
  \textit{``Monday thru Thursday is 20 trip min \& Friday thru Sunday is 40 trip min to get the bonus. Like why?''} & 14 (1,799)\\ \midrule
Concerns over inadequate payment for time and distance invested in long pickups and waits. &
  \textit{``I don’t understand what the hell is Uber thinking when giving us long pickups for short trips. NOBODY sane accepts a \$4 ride for someone half an hour away.''} & 11 (1,429)\\ \midrule
  A rise in driver numbers contributing to fewer available rides and lower earnings. &
  \textit{``Now that millions of new drivers have joined overnight due to the unemployment benefits ending, rides have dried up.''} & 5 (658) \\  \bottomrule
\end{tabular}%
}
\caption{Most representative concerns on Reddit relevant to the need for enhancing predictability and worker agency (Total = 12,728).}
\label{tab:reddit-predictability}
\end{table}
\begin{table}[htb]
\small
\resizebox{\textwidth}{!}{%
\begin{tabular}{@{}p{0.42\linewidth} |p{0.53\linewidth} | p{0.05\linewidth} @{}}
\toprule
\multicolumn{1}{c|}{\textbf{Harm}} &
  \multicolumn{1}{c|}{\textbf{Quote}} & \multicolumn{1}{c}{\textbf{\% (Count)}}\\ \midrule
Drivers experience technical difficulties with app navigation and route mapping, leading to disruptions. &
  \textit{``I started using the Lyft nav, then Google maps, then Waze only to go back to the Lyft nav. I went through that cycle because the Lyft nav sucks but at least you see everything going on with the app. With the others, I’ve lost out on streaks and pings because of glitches, etc.''} & 20 (1,208) \\ 
 \midrule
Challenges in contesting false passenger complaints and inadequate platform responses affecting driver status. &
  \textit{``They claimed the fare was adjusted due to a rider complaint or something. They’re lying.''} & 16 (990) \\ \midrule
Drivers express risks due to dangerous multitasking and unsafe areas suggested by algorithms &
  \textit{``If a Pax wants to be taken to a part of town I don’t feel safe going, I feel like I should know, for my own safety and peace of mind.''} & 15 (909) \\ \midrule
Difficulty reaching and obtaining help from support for various driver safety issues &
  \textit{``Support was really quick to respond...only took 13 days.''} & 14 (863)\\ \midrule
Drivers are not fairly paid for extensive wait times during stops or added tasks &
  \textit{``The stop system is bullshit. You make less money, 100\%. I just checked. To order two rides, one there, one back, \$17. To order the same ride, but using stops it's \$13.''} & 14 (856) \\ \bottomrule
\end{tabular}%
}
\caption{Most representative concerns on Reddit relevant to the need for better safety and more time (Total = 6,144).}
\label{tab:reddit-safety}
\end{table}
\begin{table}[htb]
\small
\resizebox{\textwidth}{!}{%
\begin{tabular}{@{}p{0.45\linewidth} |p{0.50\linewidth} |>{\centering\arraybackslash} p{0.04\linewidth} @{}}
\toprule
\multicolumn{1}{c|}{\textbf{Harm}} &
  \multicolumn{1}{c|}{\textbf{Quote}} & \multicolumn{1}{c}{\textbf{\% (Count)}}\\ \midrule
Drivers express discontent with the percentage of fares they receive compared to what the platform charges customers. &
  \textit{``Yeah it's always high ride demand and they charge the passengers high prices and drivers earnings are like 25\%''} & 24 (1,025) \\ \midrule
  Concerns over unclear or unfair deactivation decisions based on customer complaints or automated systems. &
  \textit{``I canceled two orders for the distance and my account was deactivated for fraud.''} & 20 (850)\\ \midrule
Uncertainty about how ratings are determined and their direct impact on drivers' work opportunities. &
  \textit{``I read that Lyft will try to match a pax with a driver that both had 5 starred each other in the past.''} & 9 (384) \\ \midrule
Drivers feel platform algorithms unfairly influence ride distribution, surge pricing, and overall income. &
  \textit{``a lady was standing next to my car and ordered a lyft and it gave her a driver that was outside of the Bonus zone and 20 minutes away.''} & 7 (316) \\ \midrule
  Concern over algorithms distributing rides unfairly or prioritizing certain drivers &
  \textit{``Perceived favoritism by the app's algorithm towards newer drivers can limit earnings for long-term drivers.''} & 4 (151) \\ \bottomrule 

\end{tabular}%
}
\caption{Most representative concerns on Reddit relevant to the need for ensuring fairness and non-discrimination (Total = 4,280).}
\label{tab:reddit-fairness}
\end{table}

\clearpage

\section{BERTopic Hyperparameters}
\label{asec:bertopic-hyperparams}
\begin{lstlisting}
# Load CSV data
df = pd.read_csv(filename)

# Concatenate title and description into one text column
df['text'] = df['title'] + " " + df['description']

dim_model = UMAP(n_neighbors=15,
                 n_components=10,
                 min_dist=0.0,
                 metric='cosine',
                 random_state=42)

# Initialize BERTopic
topic_model = BERTopic(
    nr_topics="auto",
    min_topic_size=100,
    n_gram_range=(1, 2),
    umap_model=dim_model
)

# Fit BERTopic on the concatenated texts from the CSV
topic_model.fit(df['text'])
\end{lstlisting}

\twocolumn
\section{Financial Expenditure of Using LLMs}
\label{asec:costs}
In this section, we delve into the costs incurred from utilizing the GPT4-Turbo LLM on Microsoft Azure to synthesize concerns from the Reddit data.  

\begin{table}[ht]
\resizebox{\columnwidth}{!}{%
\centering
\begin{tabular}{ccc}
\toprule
\textbf{Cost Category} & {\textbf{Token Quantity}} & {\textbf{Cost (USD)}} \\
\midrule
Input Token Rate & {-} & {\$0.01 per 1K tokens} \\
Output Token Rate & {-} & {\$0.03 per 1K tokens} \\
\addlinespace
Total Input Tokens & 135.12 million & \$1351.20 \\
Total Output Tokens & 10.37 million & \$311.10 \\
\addlinespace
\midrule
\textbf{Total Expenditure} & {-} & \textbf{\$1662.30} \\
\bottomrule
\end{tabular}
}
\caption{GPT4-Turbo Token Usage and Associated Costs}
\label{tab:llm_costs}
\end{table}
 
We spent a total of \$1,662.30 for using GPT4-Turbo via Microsoft Azure, which our university helped fund.

\section{IRR Measurements}
For the Classification and Prevalence stages, which associated a label with a given text input, we calculated inter-rater reliability (IRR). Three trained annotators annotated 100 randomly chosen samples each for both stages. We calculated human-human IRR and human-LLM IRR using Fleiss' Kappa. To calculate human-LLM IRR, we generated the human labels by taking the majority vote of the three annotators. If no clear majority existed, i.e., all three annotators chose different labels, we labeled the class as 'Other'.
The results were as follows: For the Classification stage, the human-human IRR was 0.59, and the human-LLM IRR was 0.54. For the Prevalence stage, the human-human IRR was 0.63, and the human-LLM IRR was 0.61. These results suggest good agreement according to the interpretation guidelines by~\citet{fleiss2013statistical}. Crucially, the human-human and human-LLM IRR values differed by less than 0.05, demonstrating LLMs can substitute human labels.

\end{document}